\documentstyle[close]{icassp97}

\title{Hybrid language processing in the Spoken Language Translator}
\author{Manny Rayner \and David Carter}
\affiliation{
SRI International\\
23 Millers Yard\\ 
Mill Lane\\
Cambridge CB2 1RQ\\
United Kingdom\\
\ \\
\tt{\{manny,dmc\}@cam.sri.com}
}

\begin{document}

\maketitle
\begin{abstract}
The paper presents an overview of the Spoken Language Translator (SLT)
system's hybrid language-processing architecture, focussing on the way
in which rule-based and statistical methods are combined to achieve
robust and efficient performance within a linguistically motivated
framework. In general, we argue that rules are desirable in order to
encode domain-independent linguistic constraints and achieve
high-quality grammatical output, while corpus-derived statistics are
needed if systems are to be efficient and robust; further, that hybrid
architectures are superior from the point of view of portability to
architectures which only make use of one type of information. We
address the topics of ``multi-engine'' strategies for robust
translation; robust bottom-up parsing using pruning and grammar
specialization; rational development of linguistic rule-sets using
balanced domain corpora; and efficient supervised training by
interactive disambiguation.  All work described is fully implemented
in the current version of the SLT-2 system.
\end{abstract}

\section{Introduction}

When attempting to implement a language-processing architecture for any
kind of practically useful speech understanding system, there is a
tension between two fundamental requirements. Other things being
equal, we would like our language processing to be based on
declarative, linguistically motivated descriptions of language; this
gives us the advantages of increased portability across domains and
(to a lesser extent) languages, and makes it easier to incorporate
insights from theoretical linguistics into the system. However,
important as these goals are, it is even more important that the
system be at least moderately fast and robust. A system which is
too slow and brittle is not of great practical interest,
even if it has a theoretically impeccable pedigree.

There is at the moment wide-spread disenchantment with the idea of
building systems based on large hand-coded grammars. Critics of
the approach generally offer some variant of an argument which can
briefly be summarized as follows:
\begin{enumerate}
\item Grammars take too long to develop.

\item They always leak badly.

\item They need substantial manual tuning to give reasonable
coverage in a new domain.

\item Even after doing that, processing is still very slow.

\item At the end of the day, performance is anyway no better than 
        what you would get from a surface processing method.

\item So why bother?
\end{enumerate}
We do not think these objections are unreasonable; they are based on
many people's painful experience, including our own. However, we do
not believe either that the problems listed above are insurmountable.
This paper gives an overview of the methodology we have developed
for attacking them. In equally brief form, our response is:
\begin{enumerate}
\item We keep our grammars, lexica and other linguistic descriptions
as general as possible, so that the large development cost is a
one-off investment.

\item We make sure that the grammar contains all, or nearly all, of the 
difficult core constructions of the language; then most coverage holes
are domain-specific, and fairly easy to locate and fix.

\item Non-trivial tuning is needed when adapting the grammar for use
in a specific domain. However, a large portion of this tuning can
be performed semi-automatically with supervised training procedures
usable by non-expert personnel; the remaining work can be organized
efficiently using balanced corpora to direct expert attention where it
will be most productive.

\item Automatic corpus-based tuning of the language description
by grammar specialization and pruning makes grammar-based language
processing acceptably efficient.

\item Bottom-up processing strategies can intelligently
combine the results of ``deep'' linguistic processing and fall-back
processing using shallow surface methods. 

\item The results show a clear improvement over those produced by the
surface methods alone.
\end{enumerate}
The work we describe here is based on the SRI Core Language Engine
(CLE) and the Spoken Language Translator (SLT), both of which have
been extensively described elsewhere. The CLE \cite{CLE} is a general
language-processing system which has been developed at SRI Cambridge
under a series of projects starting in 1986.  Versions exist for
English, Swedish \cite{GambackRayner:92}, French and Spanish
\cite{RaynerEtAl:96}; several other languages are under
development. SLT \cite{SLT-HLT,SLT-report} is a speech translation
system which uses the CLE as its language processing component. The
current version of SLT is configured to translate between English,
Swedish and French in the ATIS \cite{ATIS} domain, employing a
vocabulary of about 1500 words.

In the remainder of the paper, we will focus on the issues of
portability, speed and robustness in the context of the Spoken
Language Translator's hybrid language-processing
architecture. Section~\ref{Section:Architecture} describes the overall
design of the SLT system, and Section~\ref{Section:Robust-parsing} the
aspects concerning robust parsing with domain-specialized linguistic
descriptions. Section~\ref{Section:Adaptation} describes the
semi-automatic domain adaptation
process. Section~\ref{Section:Summary} concludes.

\section{Overall System Architecture}
\label{Section:Architecture} 

This section gives an brief overview of the SLT system's language
processing architecture. Input is in the form of a speech hypothesis
lattice derived by aligning and conflating the top five sentence
strings produced by a version of the DECIPHER (TM) recognizer
\cite{Murveit:93}. Versions of this recognizer now exist for English,
Swedish and French. The lattice is passed to a source-language copy of
the CLE, where it is analysed using a robust bottom-up parsing method
described in more detail in Section~\ref{Section:Robust-parsing}

At several points during analysis, the main flow of processing pauses
to allow extraction, using a stack-decoder algorithm, of the current
best sequence of analysis fragments. (The criteria for ``best
sequence'' include both the acoustic plausibility scores delivered by
the recognizer and the linguistic information acquired during the
analysis process).  Currently, this extraction is performed at four
levels intuitively corresponding to completion of the following
processing stages: i) receipt of raw recognizer output, ii) lexical
lookup and tagging, iii) parsing of small phrases and iv) full
parsing. As soon as one of these sequences of fragments has been
produced, it is sent over to a target language copy of the CLE, which
uses them as input to transfer and generation.

The transfer and generation process has the basic flavour of the
``multi-engine'' approach described in \cite{FrederkingNirenburg:94},
and works as follows. As each sequence of fragments is received by the
target language process, they are translated and entered into a
chart-like structure whose vertices mirror the vertices of the chart
used by the source language process. Thus the translation of a
sequence of fragments which starts at vertex $V_1$ and ends at vertex
$V_2$ will be inserted into the ``translation chart'' as an edge
connecting the same vertices $V_1$ and $V_2$.

Translation is performed using two different methods. The first is
itself a hybrid method which combines unification-based rules and
statistical preferences \cite{QLF-transfer,RaynerBouillon:95}. The
rules suggest candidate translations, which are then ordered by the
statistical preferences. The basic division of effort is that rules
encode domain-independent grammatical phenomena, while the preferences
take care of problems concerning lexical choice, which are usually to
some extent domain-dependent. The unification-based method is
only applicable to fragments produced during the last two stages
of processing, when at least some grammatical information has been
applied.

If the unification-based method is able to produce a translation at
all, it is generally of high quality. However, like most rule-based
systems, the unification-based transfer method tends to be somewhat
fragile. We consequently supplement it with a less sophisticated
surface translation method, which simply associates source-language
words and phrases (optionally tagged by part-of-speech) with
target-language equivalents. The bilingual phrasal lexicon which
encodes the relevant information is built semi-automatically using a
simple corpus-based tool.

Action of the two translation methods results in addition of
successively longer edges to the ``translation chart''; since the
methods are run starting with the simplest ones, there is a spanning
set of edges from an early stage. At any point in the process, it
is thus possible to pause and extract a current best sequence of
translation fragments from the chart. Extraction is performed using
the same stack-decoder algorithm as is used on the analysis side.

The visible result is that the translation process produces a first
rough version of the translation very quickly, using the surface
method; it then refines it over several iterations as edges produced
by the deep translation method become available. When no more edges
are available for processing, or alternately when a pre-set time-limit
has been exceeded, the best sequence of translation fragments for the
final version of the chart is extracted and sent to a speech
synthesizer.

\section{Linguistically Motivated Robust Parsing}
\label{Section:Robust-parsing}

The previous section described how grammar-based parsing contributes
to the general robust translation scheme. We now consider in more
detail the question of how a corpus can be used to specialize a
general grammar so that it delivers practically useful performance in
a given domain.

The central problem is easy to state. By the very nature of its
construction, a general grammar allows a great many theoretically
valid analyses of almost any non-trivial sentence.  However, in the
context of a specific domain, most of these will be extremely
implausible, and can in practice be ignored. To achieve efficient
parsing, we need to be able to focus our search on only a small
portion of the space of theoretically valid grammatical analyses. Our
work on parsing (see \cite{ACL96} for a fuller description, albeit one
based on a slightly earlier version of the system) is a logical
continuation of two specific strands of research aimed in this general
direction.

The first is the popular idea of {\it statistical tagging} e.g.\
\cite{DeRose,XeroxTagger,ChurchTagger}.  Here, the basic idea is that
a given small segment $S$ of the input string may have several
possible analyses; in particular, if $S$ is a single word, it may
potentially be any one of several parts of speech.  However, if a
substantial training corpus is available to provide reasonable
estimates of the relevant parameters, the immediate context
surrounding $S$ will usually make most of the locally possible
analyses of $S$ extremely implausible.  In the specific case of
part-of-speech tagging, it is well-known \cite{DeMarckenACL} that a
large proportion of the incorrect tags can be eliminated ``safely'',
i.e.\ with very low risk of eliminating correct tags. We generalize
the statistical tagging idea to a method called ``constituent
pruning''; this acts on local analyses (constituents) for phrases
normally larger than single-word units. Constituents are pruned out
if, on the basis of supervised training data (see Section
\ref{Section:Training} below), they seem unlikely to contribute to
subsequent parsing operations leading to an optimal analysis of the
full sentence.  Pruning decisions are based both on characteristics
of the constituents themselves and on the tags of neighbouring
constituents.  From each constituent and pair of neighbouring
constituents, a {\it discriminant} (abbreviated description of the
constituent or pair) is extracted, and the number of times this
constituent or pair has led to a successful parse in training is
compared to the number of times it was created
\cite{DaganItai:94,Yarowsky:94}. Constituents that are never or very
seldom successful on their own, or that only participate in similarly
unpromising pairs, are pruned out, unless this would destroy the
connectivity of the chart.

The second idea we use is that of {\it Explanation-Based Learning}
(EBL; \cite{MitchellEBL,HarmelenBundyEBL}). We extend and generalize
the line of work described in
\cite{FGCS-88,EBL-ARPA,EBL-IJCAI-91,SLT-report-EBL,Christer-ACL-94}. Here,
the basic idea is that grammar rules tend in any specific domain to
combine much more frequently in some ways than in others. Given a
sufficiently large corpus parsed by the original, general, grammar, it
is possible to identify the common combinations of grammar rules and
``chunk'' them into ``macro-rules''. The result is a ``specialized''
grammar; this has a larger number of rules, but a simpler structure,
allowing it in practice to be parsed very much more quickly using an
LR-based method \cite{Christer-COLING-94}. The coverage of the
specialized grammar is a strict subset of that of the original
grammar; thus any analysis produced by the specialized grammar is
guaranteed to be valid in the original one as well. The practical
utility of the specialized grammar is largely determined by the loss
of coverage incurred by the specialization process. We show in
\cite{ACL96} that suitable ``chunking'' criteria and a training corpus
of a few thousand utterances in practice reduce the coverage loss to a
level which does not affect the performance of the system to a
significant degree.

The two methods, constituent pruning and grammar specialization, are
combined as follows. The rules in the original, general, grammar are
divided into two sets, called {\it phrasal} and {\it non-phrasal}
respectively. Phrasal rules, the majority of which define fairly
simple noun phrase constructions, are used as they are; non-phrasal
rules are combined using EBL into chunks, forming a specialized
grammar which is then compiled further into a set of LR-tables.  Each
chunk applies at a particular level of parsing, depending on the kind
of constituent it can create.  Parsing proceeds bottom-up by
interleaving constituent creation and deletion. First, the lexicon and
morphology rules are used to hypothesize word analyses.  Constituent
pruning then removes all sufficiently unlikely edges.  Next, the
phrasal rules are applied bottom-up, to find all possible phrasal
edges, after which unlikely edges are again pruned. Finally, the
specialized grammar is used to search for constituents at successively
higher levels; pruning may be carried out after any of these levels
has been completed, the decision depending on whether pruning at a
given level offers an overall speedup for the domain in question.

\section{Semi-automatic domain adaptation of grammars}
\label{Section:Adaptation}

Section~\ref{Section:Robust-parsing} described how a general grammar
can be made to deliver useful performance, measured in terms of speed
and robustness, within a given domain. We now discuss the related
question of how to achieve acceptable coverage. Our experience is that
when an unmodified general grammar is used to process utterances from
a given domain, it fails badly in two respects.  Firstly, there are
virtually always a number of serious coverage holes, reflecting
constructions common in the domain which are inadequately handled
by the grammar. Secondly, there is the ubiquitous problem of
ambiguity; even when the coverage holes are fixed, most utterances
receive multiple analyses, of which only a small proportion are
correct. 

In the remainder of the section, we discuss these two problems.
In Section~\ref{Section:Rational-development}, we describe a
simple methodology which allows us rapidly to identify and fix the
important coverage holes in the grammatical rule sets. 
Section~\ref{Section:Training} describes a supervised training method
which attacks the problem of ambiguity.

\subsection{Rational Development of Rule Sets}
\label{Section:Rational-development}

An unmodified general grammar generally delivers poor coverage in a
specific domain. However, our experience is that most of the
utterances which fail to parse do so because of a relatively small
number of isolated problems; these are typically missing lexical
entries, missing grammar rules for idiosyncratic types of phrase
common in the domain, and minor faults in existing rules. Grammar bugs
of this kind are easy to fix. The real problem is identifying them
quickly and efficiently, so that effort is focussed on bugs which
significantly affect coverage in the given domain.

Our methodology is based on the idea of constructing a
``representative subcorpus''. By this, we mean a small
subset of the main corpus, intelligently selected so as to 
exemplify the important domain constructions in descending frequency
order. Our recipe for constructing representative subcorpora is
roughly as follows, and can be used by non-experts who have
some basic familiarity with linguistics.
\begin{enumerate}

\item Since the grammar will operate bottom-up, it is unnecessary to
be able to analyze all utterances as complete units. Thus start by
dividing long utterances into smaller pieces, which can reasonably
be thought of as units to be translated separately.

\item Assign part-of-speech tags to the ``split'' utterances using
some kind of tagger.

\item Group utterances into equivalence classes under the relationship
of having the same tag-sequence.

\item Manually regroup the classes produced by the previous step where
necessary. In some cases, this involves reclassifying utterances which
were incorrectly tagged; in others, a group may be split into two or
three smaller groups, if the relevant utterances are intuitively
dissimilar enough. This step can be performed by non-experts at the
rate of several thousand sentences a day, using a simple interactive
tool.

\item For each of the new classes, manually designate an element which
intuitively is ``most typical'' of the class. This step can also be
performed quickly by non-experts using the same interactive tool.

\item Construct the ``representative subcorpus'' by selecting
the designated element from each class. Order the results by the
size of the classes represented.

\end{enumerate}
Our experience is that by starting at the top of the representative
subcorpus and working downwards, it is possible to fix the important
coverage problems in a new corpus with an investment of only a few
weeks of expert effort. The representative subcorpus is also a valuable
resource for performing subsequent routine system testing.

\subsection{Training by Interactive Disambiguation}
\label{Section:Training}

We have already indicated how discriminants are used at run-time to
decide which constituents should be pruned. Similar discriminants are
also used to choose between alternative analyses for a sentence.
However, deriving discriminant statistics involves selecting the
correct analysis. This requires human intervention, and we would
prefer the human in question not to have to be a system expert; but
even for an expert, inspecting all the analyses for every sentence
would be a tedious and time-consuming task. There may be dozens of
quite detailed analyses that are variations on a small number of
largely independent themes: choices of word sense, modifier
attachment, conjunction scope and so on.

It turns out that some kinds of discriminant can be presented to
non-expert users in a form they can easily understand. For training on
an utterance to be effective, we need to provide enough such
``user-friendly'' discriminants to allow the user to select the
correct analyses, and as many as possible ``system-friendly''
discriminants that, over the corpus as a whole, distinguish reliably
between correct and incorrect analyses and can be used for this
purpose at run time, either in constituent pruning or in preferring
one analysis from a competing set. Ideally, a discriminant will be
both user-friendly and system-friendly, but this is not essential.

We have developed an interactive program, the TreeBanker
\cite{CarterKeegan}, which maintains a database of the discriminants
that apply to the different analyses of each sentence in a corpus. It
presents discriminants to the user in a convenient graphical
form. Among the most useful discriminants are the major categories for
possible constituents of a parse; thus for the sentence ``Show me the
flights to Boston'' the string ``the flights to Boston'' as a noun
phrase discriminates between the correct reading (with ``to Boston''
attaching to ``flights'') and the incorrect one (with it attaching to
``show'').  Other discriminants describe semantic triples of head,
modifier and dependent (for example, ``flight+to+Boston'', which is
correct, and ``show+to+Boston'', which is incorrect), and other
information about analyses such as the sentence type or mood.

The user may click on any discriminant to select it as correct or
incorrect.  Typically there will be far more discriminants presented
than the number of distinct differences between the analyses.  The
effect of this is that users can give attention to whatever
discriminants they find it easiest to judge; other, harder ones will
typically be resolved automatically by the TreeBanker as it reasons
about what combinations of discriminants apply to which analyses.  For
example, when the CLE analyses the sentence ``What is the earliest
flight that has no stops from Washington to San Francisco on
Friday?'', it yields 154 analyses and 318 discriminants, yet the
correct analysis may be obtained with only two selections.  Selecting
``the earliest flight ... on Friday'' as a noun phrase eliminates all
but twenty of the analyses produced, and approving ``that has no
stops'' as a relative clause eliminates eighteen of these, leaving
analyses which are both correct for the purposes of translation.  152
incorrect analyses may thus be dismissed in a few seconds.

\section{Summary}
\label{Section:Summary}

We have described a methodology which, in our opinion, demonstrates
that hybrid approaches based on general hand-coded grammars can be
practically useful in the context of a realistic speech-understanding
task like medium-vocabulary spoken language translation. In
particular, we have addressed what we see as the key questions:
achieving adequate speed, robustness and coverage within a specific
domain, and adapting the general grammar to the domain without
excessive effort.

We regret that space restrictions make it impossible for us to present
detailed performance results here. We refer interested readers to the
SLT-2 final report, which by the time of the conference will be
available over WWW from {\tt http://www.cam.sri.com}.

\section*{Acknowledgements}

Work on the generic and Swedish-specific aspects of the Spoken
Language Translator project is funded by Telia Research AB, Stockholm,
Sweden. French-specific work has been mainly funded by SRI
International and Suissetra.

This paper appears in ICASSP-97 and is \copyright\ IEEE, 1997.

\end{document}